\documentclass[11pt, a4paper, logo, copyright, nonumbering]{SAI}
\usepackage[authoryear, sort&compress, round]{natbib}
\usepackage{dblfloatfix}
\usepackage{ulem}
\usepackage{caption}
\usepackage{dramatist}
\usepackage{xspace}
\usepackage{pifont} %
\usepackage{multirow}
\usepackage{tcolorbox}
\usepackage{xltabular}
\usepackage{longtable}
\usepackage{hyperref}
\usepackage{tcolorbox}
\usepackage{tablefootnote}
\usepackage{amsfonts}
\usepackage{amsmath}
\usepackage{amssymb}
\usepackage{lineno}
\usepackage{multirow}
\usepackage{adjustbox}

\usepackage[bottom]{footmisc}

\usepackage{CJKutf8}
\usepackage{subfigure}
\usepackage{setspace}

\usepackage{dsfont}
\usepackage{array} %
\usepackage{tabularx} %
\usepackage{subfigure} %
\usepackage{xcolor} %
\usepackage{tabularx}
\usepackage{booktabs}

\usepackage{lipsum}  %
\usepackage{multicol} %

\makeatletter
\def\@BTrule[#1]{%
  \ifx\longtable\undefined
    \let\@BTswitch\@BTnormal
  \else\ifx\hline\LT@hline
    \nobreak
    \let\@BTswitch\@BLTrule
  \else
     \let\@BTswitch\@BTnormal
  \fi\fi
  \global\@thisrulewidth=#1\relax
  \ifnum\@thisruleclass=\tw@\vskip\@aboverulesep\else
  \ifnum\@lastruleclass=\z@\vskip\@aboverulesep\else
  \ifnum\@lastruleclass=\@ne\vskip\doublerulesep\fi\fi\fi
  \@BTswitch}
\makeatother

\addto\extrasenglish{
}

 {\begin{list}{}%
         {\setlength{\leftmargin}{#1}}%
         \item[]%
 }
 {\end{list}}
 
\reportnumber{001} %

\title{\textsc{PRL-Bench}: A Comprehensive Benchmark for Evaluating the Capabilities of LLMs in Frontier Physics Research}
\author{
    Tingjia Miao\textsuperscript{\rm 1,2,3,9} \quad
    Wenkai Jin\textsuperscript{\rm 1} \quad
    Muhua Zhang\textsuperscript{\rm 3,4,5} \quad
    Jinxin Tan\textsuperscript{\rm 3,4,5} \quad 
    Yuelin Hu\textsuperscript{\rm 2,3} \quad
    Tu Guo\textsuperscript{\rm 3,4} \quad
    Jiejun Zhang\textsuperscript{\rm 3,4} \quad
    Yuhan Wang\textsuperscript{\rm 2,3,6} \quad
    Wenbo Li\textsuperscript{\rm 2} \quad
    Yinuo Gao\textsuperscript{\rm 2,3} \quad
    Shuo Chen\textsuperscript{\rm 7} \quad
    Weiqi Jiang\textsuperscript{\rm 7} \quad
    Yayun Hu\textsuperscript{\rm 8} \quad
    Zixing Lei\textsuperscript{\rm 1} \quad
    Xianghe Pang\textsuperscript{\rm 1,9} \quad
    Zexi Liu\textsuperscript{\rm 1,9} \quad
    Yuzhi Zhang\textsuperscript{\rm 9} \quad
    Linfeng Zhang\textsuperscript{\rm 10} \quad
    Kun Chen\textsuperscript{\rm 7} \quad
    Wei Wang\textsuperscript{\rm 3,4,5} \quad
    Weinan E\textsuperscript{\rm 1} \quad
    Siheng Chen\textsuperscript{\rm 1,9}
    \\
    \textsuperscript{\rm 1} School of Artificial Intelligence, Shanghai Jiao Tong University \quad \\
    \textsuperscript{\rm 2} Zhiyuan College, Shanghai Jiao Tong University\quad  \\
    \textsuperscript{\rm 3} School of Physics and Astronomy, Shanghai Jiao Tong University \quad \\
    \textsuperscript{\rm 4} Tsung-Dao Lee Institute, Shanghai Jiao Tong University\quad \\
    \textsuperscript{\rm 5} State Key Laboratory of Dark Matter Physics, Shanghai Jiao Tong University  \quad \\
    \textsuperscript{\rm 6} Shanghai Innovation Institute\quad  \\
    \textsuperscript{\rm 7} Institute of Theoretical Physics, Chinese Academy of Sciences\quad  \\
    \textsuperscript{\rm 8} Zhejiang Lab \qquad 
    \textsuperscript{\rm 9} SciLand \quad 
    \textsuperscript{\rm 10} DP Technology
    }

\newcommand{\xmark}{\textcolor{red}{\ensuremath{\times}}}

\begin{abstract}
The paradigm of agentic science requires AI systems to conduct robust reasoning and engage in long-horizon, autonomous exploration. However, current scientific benchmarks remain confined to domain knowledge comprehension and complex reasoning, failing to evaluate the exploratory nature and procedural complexity of real-world research. In this work, we present research-oriented evaluations in theoretical and computational physics, a natural testbed with comprehensive domain knowledge, complex reasoning, and verifiable end-to-end workflows without reliance on experiments. Here we introduce \textsc{PRL-Bench} (Physics Research by LLMs), a benchmark designed to systematically map the capability boundaries of LLMs in executing end-to-end physics research. Constructed from 100 curated papers from the latest issues of Physical Review Letters since August 2025 and validated by domain experts, \textsc{PRL-Bench} covers five major theory- and computation-intensive subfields of modern physics: astrophysics, condensed matter physics, high-energy physics, quantum information, and statistical physics. Each task in the benchmark is designed to replicate the core properties of authentic scientific research, including exploration-oriented formulation, long-horizon workflows, and objective verifiability, thereby reconstructing the essential reasoning processes and research workflows of real physics research. Evaluation across frontier models and further analysis show that (i) even the strongest models achieve overall scores well below 50; (ii) failures are dominated by conceptual and formulaic errors, suggesting that domain knowledge in advanced theoretical physics remains scarce; (iii) exploration and derivations can be unstable, reflecting limitations in maintaining coherent reasoning chains over extended horizons. Thus, \textsc{PRL-Bench} can serve serve a reliable testbed for accessing next generation AI scientists advancing AI systems toward autonomous physics research. The data is available at \url{https://huggingface.co/datasets/AdrianMiao/PRL_Bench}.
\end{abstract}

\begin{document}

\maketitle

\newpage

\begin{spacing}{0.9}
\tableofcontents
\end{spacing}

\newpage

\section{Introduction}
\label{sec:introduction}

Artificial Intelligence for Science (AI4Science) is attracting increasing attention across both academia and industry. Beyond its conventional role as a scientific tool, recent advances in large language models (LLMs) and agentic systems indicate that AI4Science is entering a new phase of agentic science: shifting from assisting and accelerating isolated scientific subtasks to automating end-to-end scientific research workflows. This transition naturally raises a more fundamental question: beyond serving as tools for individual steps of scientific work, to what extent can AI function as autonomous scientific researchers?

Existing evaluations fail to adequately capture the capability requirements of agentic science. Recent benchmarks have significantly elevated the difficulty of reasoning and domain knowledge, including Olympiad-style evaluations such as OlympiadBench~\citep{he2024olympiadbench}, OlympicArena~\citep{huang2025olympicarena}, and OlymMath~\citep{sun2025olymmath}, with Humanity's Last Exam (HLE)~\citep{phan2025humanity} standing out for its breadth and rigor. Nevertheless, these benchmarks remain confined to well-defined problem settings with explicit objectives, clear solution pathways, and predetermined reasoning processes. As such, while they demonstrate LLMs' domain knowledge and general reasoning capacity, they provide limited insight into the systems' ability to autonomously plan, adapt, and explore in realistic scientific research. 

To fill this critical gap, we begin with theoretical and computational physics, an ideal domain with rigorous reasoning, usage of heterogeneous tools and verifiable end-to-end research workflows, enabling verification without reliance on experiments. However, existing physics-specific benchmarks, including TPBench~\citep{chung2025theoretical} and PHYSICS~\citep{feng2025physics}, also rely on short, clear-path tasks, failing to reflect the long-horizon and exploratory nature of authentic physics research. Frontier Science~\citep{wang2026frontierscience} advances research-oriented evaluation but features a limited-scale physics component—only 20 questions—with insufficient coverage of frontier subfields like condensed matter physics and high-energy physics.

Here we introduce \textsc{PRL-Bench} (Physics Research by LLMs), a \textbf{frontier expert-level} benchmark constructed from \textbf{authoritative sources} and of \textbf{considerable scale}, designed to systematically and objectively assess the capability boundaries of large language models in real physics research. \textsc{PRL-Bench} is constructed from \textbf{100 authoritative papers} curated from the renowned journal \textbf{Physical Review Letters} since August 2025 (volume 135 issue 7), covering astrophysics, condensed matter physics, high-energy physics, quantum information, and statistical physics. In collaboration with over ten domain experts, each paper is converted into a research-oriented task focused on reasoning and computation, possessing open solution pathways and a long-horizon structure. All tasks have passed expert cross-validation to ensure consistency with the underlying physics of the source papers.

\begin{table}[htbp]
\centering
\small
\begin{tabular}{lcccc}
\toprule
Benchmark & Scale & Knowledge Level & Research-Oriented & Physics-Specialized \\
\midrule
OlympicArena & Large (11163) & Undergraduate & \xmark & \xmark \\
HLE & Large (2500) & Graduate & \xmark & \xmark \\
PHYSICS & Large (1,297) & Undergraduate & \xmark & \checkmark \\
PHYBench & Large (500) & Graduate & \xmark & \checkmark \\
PRBench & Small (20) & Frontier Research & \checkmark & \checkmark \\
FrontierScience\tablefootnote{Denotes the research partition of the FrontierScience benchmark.} & Small (60) & Frontier Research & \checkmark & \xmark \\
\textsc{PRL-Bench} & Medium (100) & Frontier Research & \checkmark & \checkmark \\
\bottomrule
\end{tabular}
\caption{Comparison of representative benchmarks across scale, knowledge level, research orientation, and physics specialization.}
\label{tab:benchmark_comparison}
\end{table}

Building on \textsc{PRL-Bench}, we conduct a systematic evaluation of frontier large language models. We study their performance across the five physics subfields and analyze their capabilities in research-oriented tasks involving reasonable planning, long-horizon exploration, rigorous reasoning and computation.Evaluation across frontier models shows that (i) even the strongest models achieve overall scores well below 50; (ii) failures are dominated by conceptual and formulaic errors, suggesting that domain knowledge in advanced theoretical physics remains scarce; (iii) exploration and derivations can be unstable, reflecting limitations in maintaining coherent reasoning chains over extended horizons. Our analysis reveals that the combination of lack in domain knowledge, derivation stability, numerical reliability, and, critically, long-horizon task adaptation, resulting in the substantial gap between current LLM capabilities and the requirements of real physics research. Thus, \textsc{PRL-Bench} can serve a reliable testbed for accessing next generation AI scientists advancing AI systems toward autonomous physics research.

\section{Related Work}

\subsection{Scientific Benchmarks for LLMs}
Early general science benchmarks for LLMs focused on closed-ended QA to assess domain knowledge comprehension, such as ScienceQA \citep{saikh2022scienceqa} and SciBench \citep{wang2023scibench}.
work has increasingly centered on incorporating more complex reasoning and advanced domain knowledge, including various Olympiad-level tasks \citep{he2024olympiadbench,huang2025olympicarena,sun2025olymmath}. Humanity’s Last Exam (HLE) \citep{phan2025humanity} is representative in terms of difficulty and comprehensiveness but still lacks the exploratory nature of real scientific research.

OpenAI’s Frontier Science \citep{wang2026frontierscience} pioneered a new paradigm of research-oriented evaluation, but its physics-related tasks are small in scale with only 20 questions, failing to effectively cover frontier subfields such as condensed matter physics and high-energy physics.

Physics-specific benchmarks include TPBench~\citep{chung2025theoretical}, PHYSICS~\citep{feng2025physics}, PHYBench~\citep{qiu2025phybench}, which rely on short, clear-path tasks and fail to capture the long-horizon nature of authentic research. PRBench~\citep{qiu2026prbench}, a recent work sharing a similar motivation with \textsc{PRL-Bench}, focuses on end-to-end paper reproduction in physics research, with a focus on the ability to faithfully reproduce all detailed implementations and results of original studies, while \textsc{PRL-Bench} centers on the reproduction of exploratory behaviors in research and further revises tasks to increase difficulty, leading to distinct task design philosophies.

\subsection{AI for science and AI scientists}
AI has long been recognized as a transformative force in scientific research, with its role evolving from a scientific tool toward a potentially autonomous scientist. Historically, AI has been primarily applied to isolated sub-tasks rather than full research workflows. With recent advances in capability, general-purpose AI scientist systems have begun to emerge, including Google DeepMind’s AI co-scientist~\citep{natarajan2025aicoscientist}, Robin~\citep{ghareeb2025robin}, and Kosmos~\citep{mitchener2025kosmos}.
In the domain of physics, both general and specialized AI physicist systems—such as PhysMaster~\citep{miao2025physmaster}, GRACE~\citep{hill2026grace}, and ColliderAgent~\citep{qiu2026end}—are only beginning to appear. This nascent stage underscores the need for a rigorous, comprehensive, and objective evaluation framework, such as \textsc{PRL-Bench}, to assess the ability of large language models to carry out end-to-end research tasks in physics.

\section{Benchmark}

\subsection{Source}
A total of 100 authoritative papers are curated from \emph{Physical Review Letters}, spanning issues from Volume 135, Issue 7 (August 2025) to Volume 136, Issue 10 (Mar. 2026), as the source corpus. All selected papers are centered on theoretical derivation and numerical computation; works primarily focused on experimental studies, as well as those requiring large-scale datasets, substantial computational resources, or specialized simulation software, are systematically excluded.

\subsection{Subfields}

Our \textsc{PRL-Bench} spans five major subfields of modern physics:

\begin{enumerate}
\item \textbf{Astrophysics (Astro)}: Black holes and black-hole thermodynamics, compact astrophysical objects such as neutron stars and white dwarfs, gravitational-wave sources, early-universe cosmology, dark matter and dark-sector phenomenology, etc.

\item \textbf{Condensed matter physics (Cond-Mat)}: Quantum many-body systems in material settings, strongly correlated electron systems, topological phases of matter, superconductivity and superfluidity, etc.

\item \textbf{High-energy physics (HEP)}: Quantum field theory and gauge theory, QCD and non-perturbative dynamics, effective field theory, conformal field theory, theories beyond the Standard Model, etc.

\item \textbf{Quantum information and foundations (Quantum)}: Quantum error correction, tensor-network methods for quantum states, open quantum systems, quantum resource theory, and foundational aspects of quantum mechanics.

\item \textbf{Statistical physics and complex systems (Stat)}: Equilibrium and non-equilibrium statistical mechanics, stochastic processes, disordered systems, many-body dynamics, etc.
\end{enumerate}

Together, these five areas cover physical phenomena across a wide range of scales, from cosmological structures to the microscopic quantum regime in which Quantum Chromodynamics (QCD) governs the dynamics of quarks and gluons. Each area has distinct methodological characteristics: some rely primarily on formal analytic structures, others on model construction and asymptotic reasoning, and still others on effective formulations, phenomenological description, or numerical computation.

As a result, \textsc{PRL-Bench} evaluates not only whether a model can conduct robust reasoning, but also whether it can flexibly adapt appropriate methodological strategy across major physical subfields in end-to-end physics research.

\begin{figure}[ht]
\centering
\begin{minipage}[t]{0.4\textwidth}
    \centering
    \includegraphics[width=\textwidth]{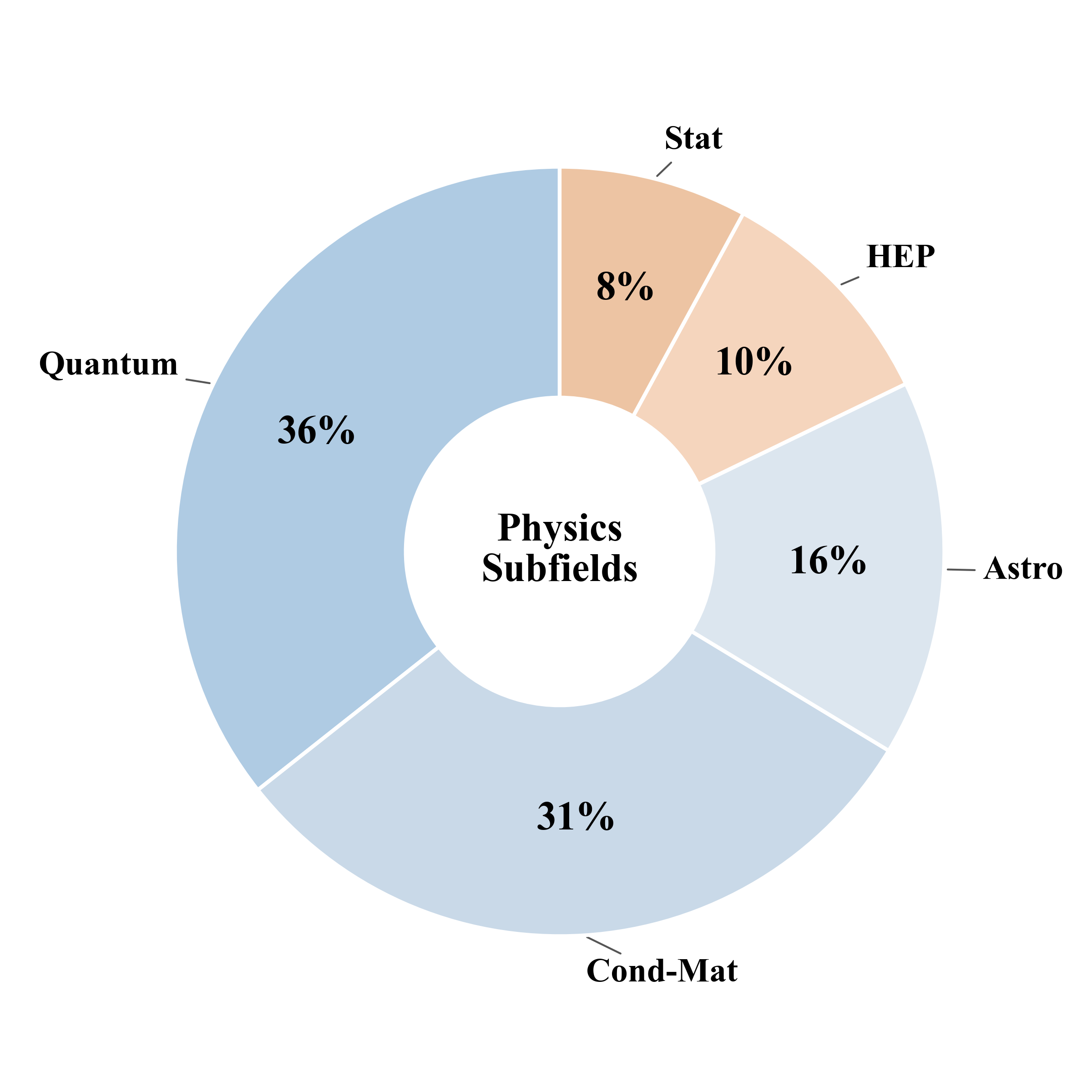}
\end{minipage}
\hfill
\begin{minipage}[t]{0.48\textwidth}
    \centering
    \includegraphics[width=\textwidth]{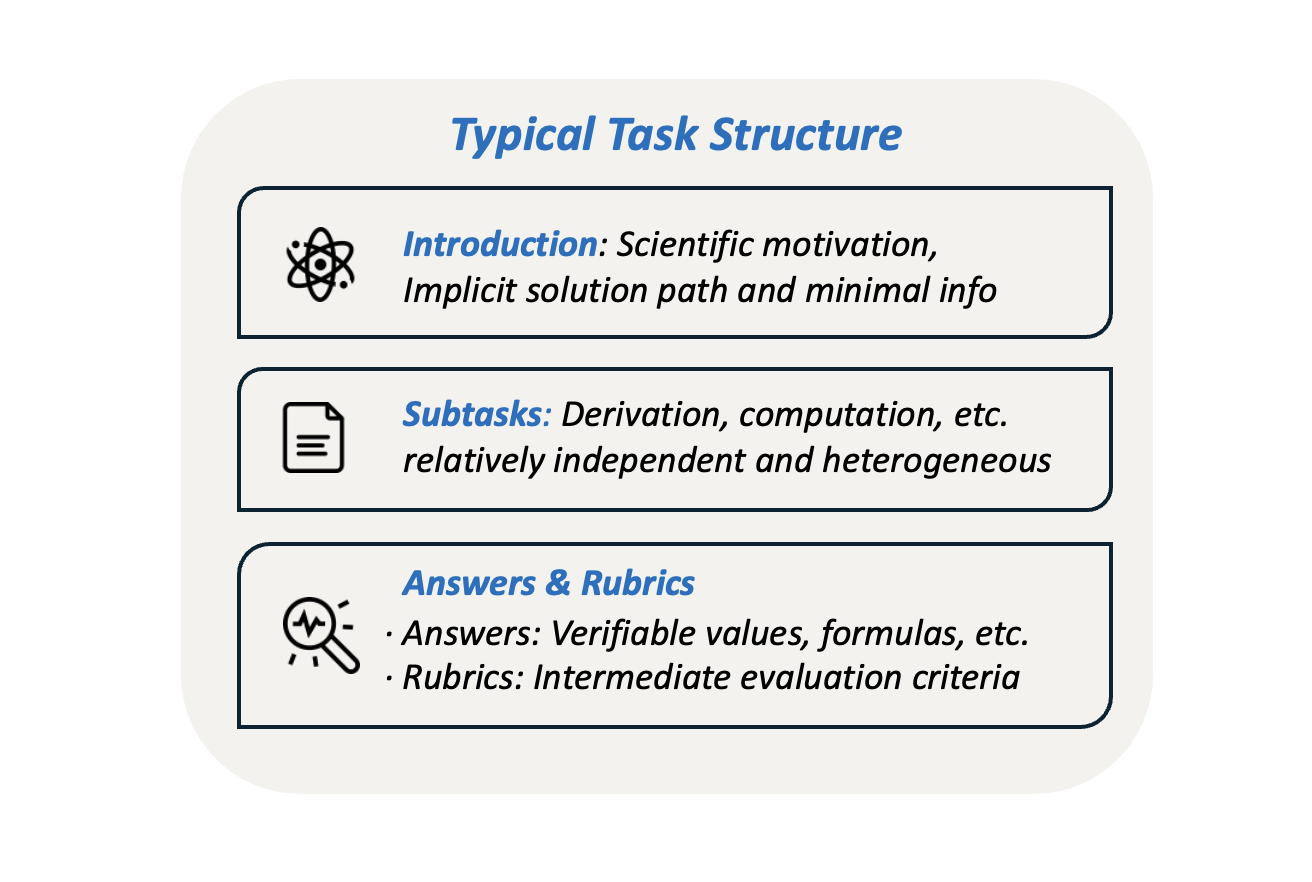}
\end{minipage}

\caption{Overview of \textsc{PRL-Bench}: \\(a) Subfield distribution of \textsc{PRL-Bench}, (b) Typical task structure of \textsc{PRL-Bench}}
\label{fig:your_label}
\end{figure}

\subsection{Task Design}

The tasks in \textsc{PRL-Bench} are designed to align the characteristics of authentic theoretical and computational physics research, rather than the closed-form single-path problems. In particular, each task is constructed to preserve three core properties of real scientific inquiry: exploration-oriented formulation, long-horizon workflows, and objective verifiability.

The core principle of task design is exploration-oriented, aiming to jointly evaluate models’ capabilities in autonomous planning, information integration, and long-horizon reasoning under conditions where solution pathways are not explicitly specified. Therefore, each task aligns with authentic research by providing a scientific motivation and a concrete research objective, while the solution pathway are implicit and domain knowledge are not stated explicitly, preserving minimal information to ensure a unique and verifiable solution.

This setting reflects real scientific inquiry, in which progress requires selecting appropriate theoretical framework, pursuing intermediate results, and iteratively refining the approach. Although the objective is precisely defined, the solution pathways must be actively determined, requiring context-sensitive deployment of domain knowledge rather than reliance on solution.

\begin{figure}[htbp]
    \centering
    \includegraphics[width=0.9\linewidth]{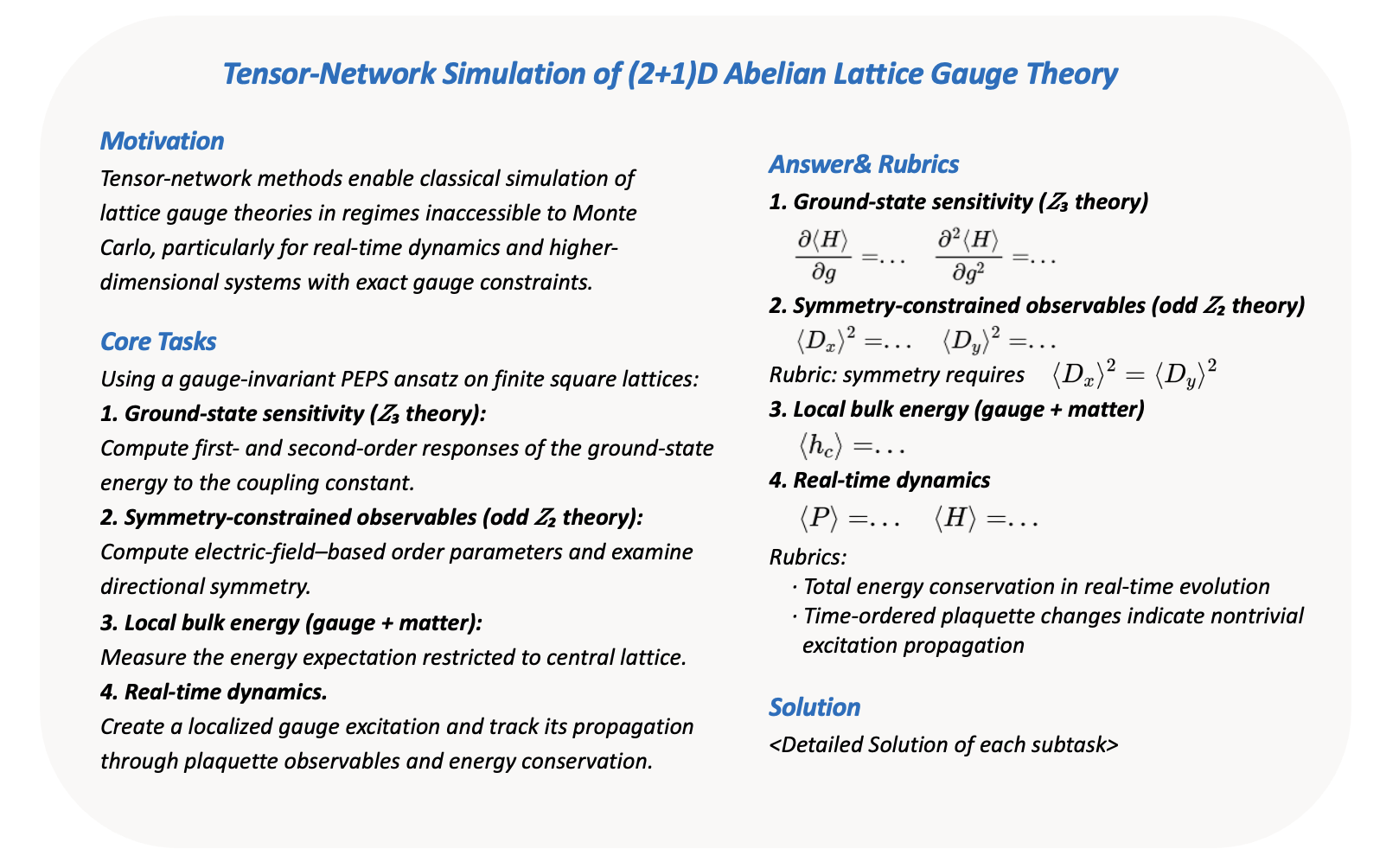}
    \caption{An representative task from \textsc{PRL-Bench}: \\Tensor-Network Simulation of (2+1)D Abelian Lattice Gauge Theory
}
    \label{example}
\end{figure}

A representative task from \textsc{PRL-Bench} is illustrated in Figure~\ref{example}, comprising four core components: motivation, core task, answers \& rubrics, and a detailed solution.

\newpage
Each task is structured as a sequence of relatively independent and heterogeneous subtasks, such as analytical derivation and computational validation. While these subtasks are unified under a shared scientific objective, they avoid forming a strictly linear dependency chain, thereby mitigating error propagation and enabling a more reliable assessment of models’ capability boundaries.

For evaluation, each subtask in \textsc{PRL-Bench} preserves both answers and rubrics: 

\begin{enumerate}
    \item \textbf{Answers}: verifiable numerical values, analytical formulas, or discrete judgments, ensuring reproducibility and objectivity despite the openness of the reasoning process.
    \item \textbf{Rubrics}: structured intermediate evaluation criteria that decompose each subtask into key reasoning steps and checkpoints, providing further insight into intermediate reasoning and failure modes and enableing more fine-grid assessment of LLMs' capabilities in conducting long-horizon scientific exploration.
\end{enumerate}

\section{Evaluation}

\subsection{Experimental Setup}

We evaluate six frontier large language models: GPT-5.4, Gemini-3.1-Pro, Claude-Opus-4.6, Doubao-Seed-2.0-Pro, Qwen-3.5-Plus, and Kimi-K2.5. All models are tested under unified prompting and tool-use setting to ensure comparability.

\textbf{Tools.} During evaluation, models are provided with access to a code interpreter, enabling numerical computation and programmatic validation when required by the task. To prevent information leakage caused by retrieving original texts and ensure the accuracy and impartiality of the evaluation, search-related tools are disabled throughout the assessment process.

\textbf{Evaluation Metric.} Each problem is independently executed five times per model to reduce stochastic variance, and results are averaged. Evaluation is conducted using an LLM-as-judge paradigm (GPT-5 as judge), which strictly verifies both (i) the correctness of final answers and (ii) whether intermediate results match the rubrics.

\textbf{Scoring.} Scores are assigned to answers and rubrics in advance. The judge model will give the final score strictly based on rubric satisfaction and answer correctness, summed across subtasks, and normalized to a 0--100 scale for reporting.

\subsection{Result}

The results indicate that even frontier models achieve overall scores well below 50 (with the best performance at 44.27), highlighting the substantial difficulty of \textsc{PRL-Bench} and its effectiveness in probing the limits of current LLMs in realistic research settings. This gap suggests that long-horizon scientific reasoning—particularly involving multi-step derivation, numerical validation, and autonomous planning—remains a major bottleneck.

\begin{table}[htbp]
  \centering
  \begin{tabular}{lcccccc}
    \toprule
    Model & Astro & Cond-Mat & HEP & Quantum & Stat & Global \\
    \midrule
    GPT-5.4              & 35.02 & 37.49 & 30.99 & 40.37 & 33.88 & 37.38 \\
    Gemini-3.1-Pro       & 37.41 & 43.74 & 47.52 & 47.64 & 40.76 & 44.27 \\
    Claude-Opus-4.6      & 28.75 & 39.36 & 40.46 & 39.98 & 32.10 & 37.40 \\
    Doubao-Seed-2.0-Pro  & 28.76 & 40.49 & 35.55 & 42.67 & 24.94 & 37.83 \\
    Qwen-3.5-Plus        & 34.51 & 42.82 & 37.16 & 43.72 & 25.87 & 40.05 \\
    Kimi-K2.5            & 27.86 & 34.42 & 31.82 & 38.16 & 25.71 & 33.89 \\
    \bottomrule
  \end{tabular}
  \caption{Average score of state-of-the-art LLMs on \textsc{PRL-Bench}, normalized to a 0--100 scale.}
\end{table}

\begin{figure}[htbp]
    \centering
    \includegraphics[width=0.8\linewidth]{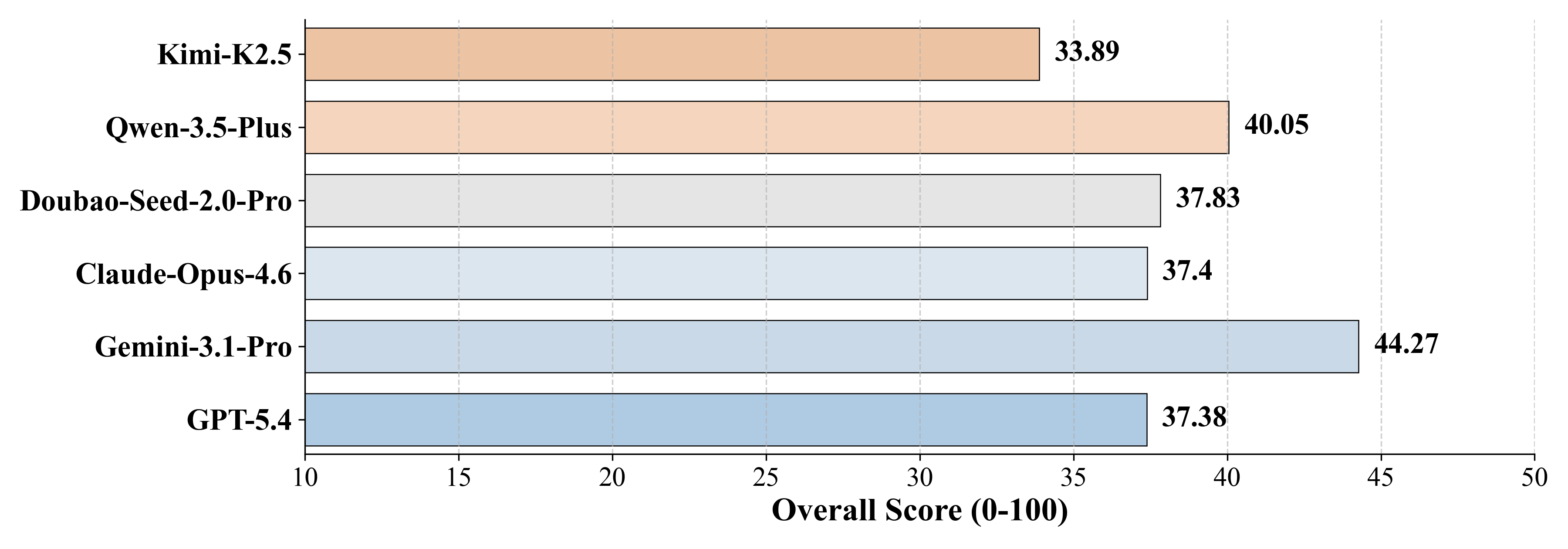}
    \caption{Average score of state-of-the-art LLMs on \textsc{PRL-Bench}}
    \label{fig:models}
\end{figure}

\newpage
As shown in Figure \ref{fig:models}, across models, Gemini 3.1 Pro consistently attains the strongest performance, achieving the highest overall score and leading in multiple subfields, indicating comparatively stronger capability in integrating heterogeneous reasoning components. Qwen 3.5 Plus ranks second, with competitive performance in Condensed Matter and Quantum domains. GPT-5.4, Claude Opus 4.6, and Doubao Seed 2.0 Pro exhibit broadly comparable performance, forming a middle tier with no clear dominance, while Kimi-K2.5 trails behind. Notably, the overall performance gap among leading models remains moderate, suggesting that current systems share similar structural limitations when confronted with research-level tasks.

\subsection{Discussion}

From the perspective of subfields, Gemini-3.1-Pro and GPT-5.4 demonstrate relatively balanced performance across domains, whereas other models exhibit more pronounced variability. In particular, most models show degraded performance in Astrophysics and Statistical Physics compared to Condensed Matter, High-Energy Physics, and Quantum Information. We infer that problems in Astro and Stat are often more heterogeneous and less standardized, resulting in weaker coverage by canonical training data and fewer reusable reasoning templates, and more comprehensive and diverse training can partially mitigate this effect.

\begin{figure}[htbp]
    \centering
    \includegraphics[width=1\linewidth]{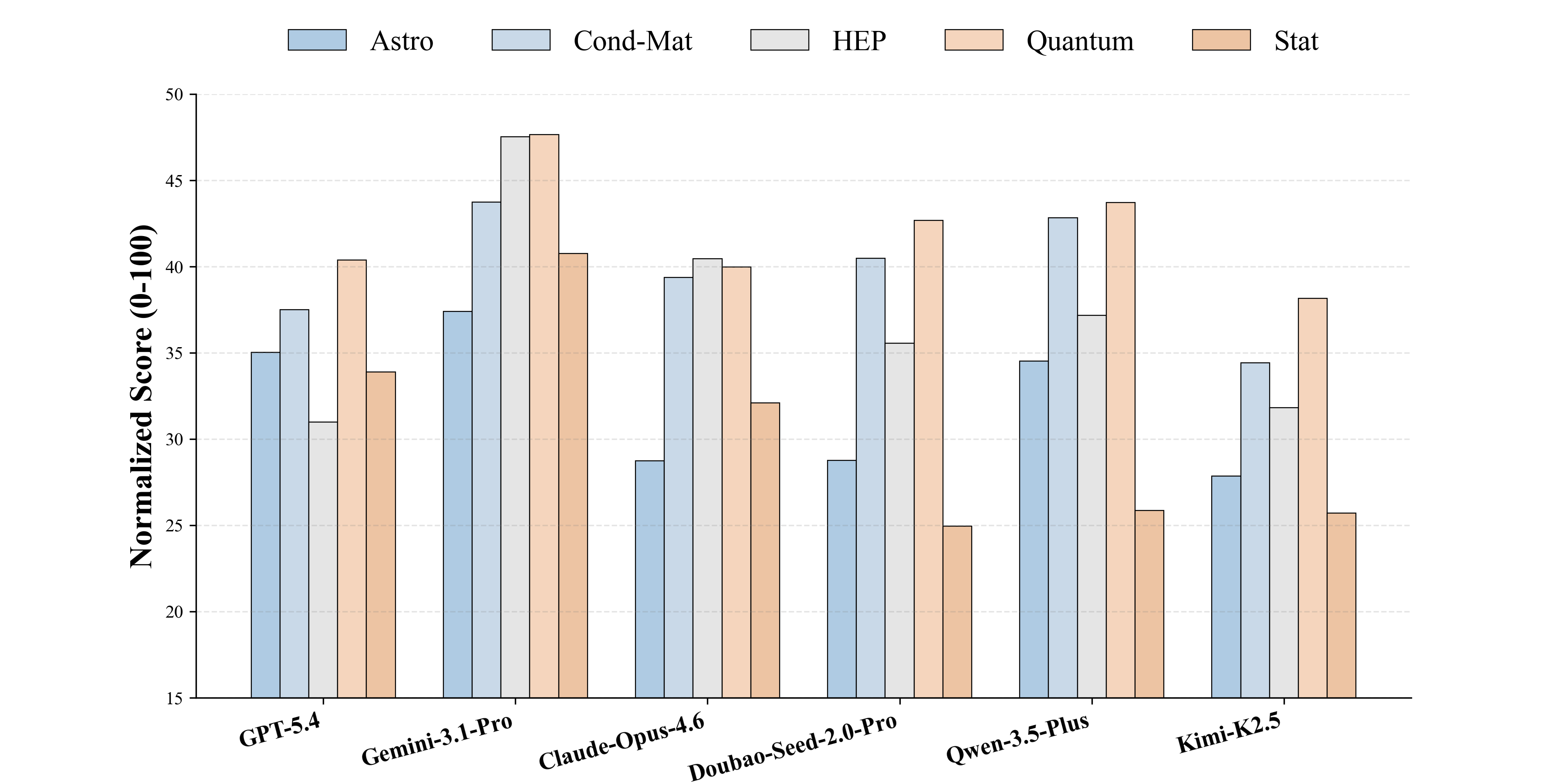}
    \caption{Average score of state-of-the-art LLMs on \textsc{PRL-Bench} across subfields}
    \label{fig:fields}
\end{figure}

\newpage
Further, we analyzed the full response trajectories of each model across different subfields and categorized the errors into four types:

\begin{enumerate}
    \item \textbf{Formulaic or conceptual error}: inappropriate choice of theoretical models or formulas, primarily reflecting insufficient domain knowledge in physics.
    \item \textbf{Derivation error}: errors arising within the derivation chain, including the use of spurious formulas or the introduction of unjustified and fabricated assumptions, primarily reflecting deficiencies in reasoning ability as well as hallucination issues.
    \item \textbf{Calculation error}: algebraic or numerical mistakes, reflecting limitations in numerical reasoning and code-based computation.
    \item \textbf{Incomplete}: omitted answers, partial answers, or failure to satisfy task requirements, primarily reflecting insufficient adaptation to long-horizon tasks, such as limitations in context management.
\end{enumerate}

\begin{figure}[htbp]
    \centering
    \includegraphics[width=1\linewidth]{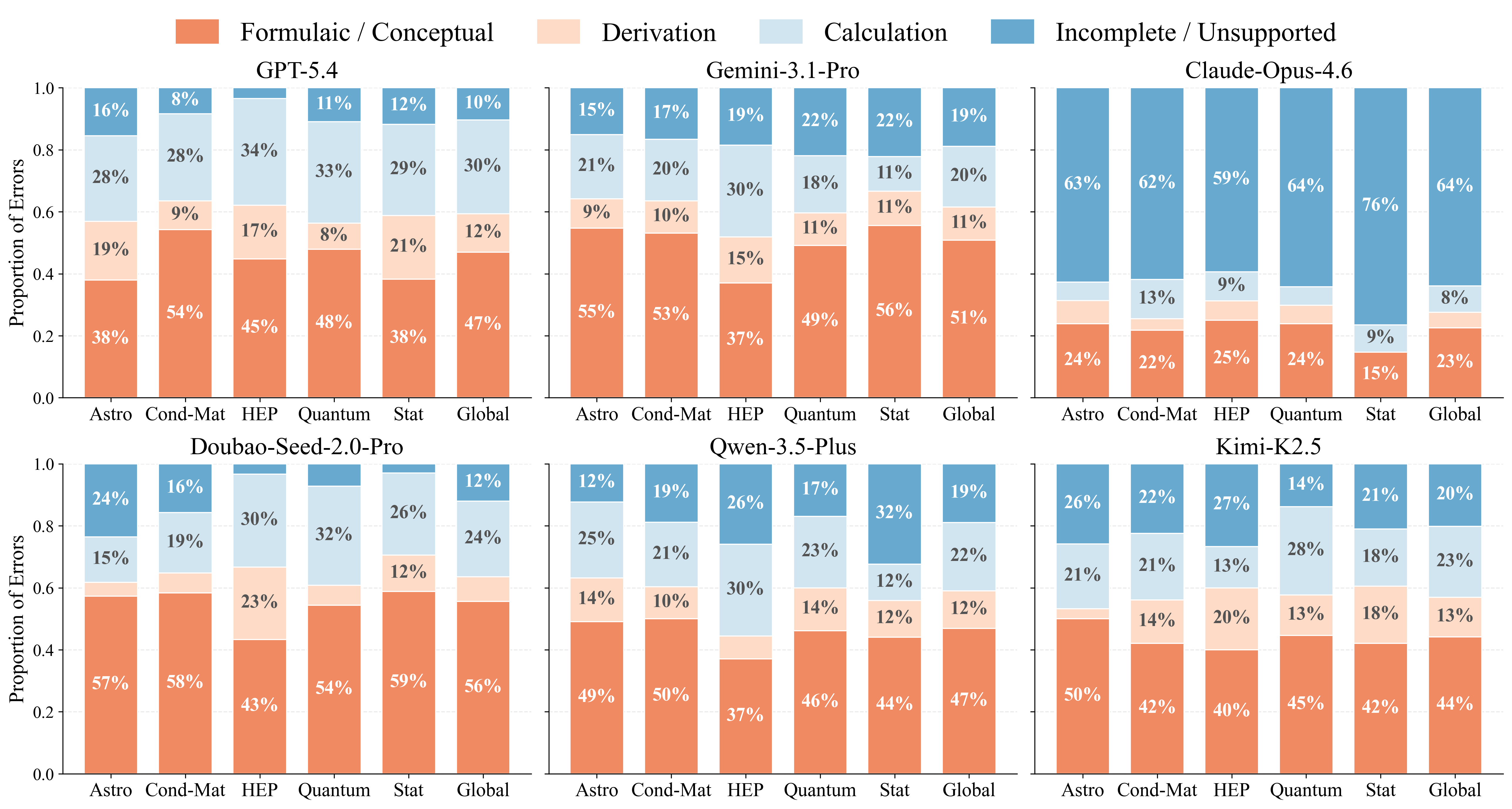}
    \caption{Error type decomposition of LLMs on \textsc{PRL-Bench} across subfields}
    \label{fig:errors}
\end{figure}

Demonstrated in Figure \ref{fig:errors}, the error type decomposition reveals several consistent patterns across models. First, formulaic or conceptual errors constitute the dominant failure mode for most models, accounting for roughly 45--55\% of errors at the global level (e.g., 0.4697 for GPT-5.4, 0.5079 for Gemini-3.1-Pro, and 0.5562 for Doubao-Seed-2.0-Pro). This indicates that incorrect or improper selection of physical models and formulas remains the primary bottleneck, even for frontier systems. The effect is particularly pronounced in domains such as Condensed Matter, where models often rely on partially matching but ultimately inappropriate theoretical templates.

\newpage
Second, derivation errors and calculation errors play more secondary but distinct roles. Derivation errors typically remain at a moderate level ($\approx 0.08$--$0.13$ globally for most models), yet become more prominent in theory-intensive domains such as HEP (e.g., 0.1724 for GPT-5.4 and 0.2333 for Doubao), reflecting instability in multi-step symbolic reasoning and a tendency to introduce invalid intermediate steps. In contrast, calculation errors are relatively stable ($\approx 0.20$--$0.30$ for most models), suggesting that algebraic manipulation and numerical computation are non-trivial but not the dominant limitation.

A distinct failure pattern is observed for Claude-Opus-4.6, where incomplete or unsupported responses dominate across all subfields (0.6393 globally). This behavior does not merely reflect conservative abstention, but is often associated with unstable long-horizon reasoning trajectories. In many failed cases, the model exhibits repeated derivation attempts and iterative self-corrections, during which unsupported assumptions are introduced to maintain superficial logical consistency. Such patterns ultimately lead to breakdowns in the research chain, resulting in incomplete or unsupported final answers. This phenomenon reflects a coupled limitation of domain knowledge, reasoning stability, and long-horizon task adaptation. Among these factors, insufficient adaptation to long-horizon tasks appears to be the primary driver, as the model lacks strategic planning and coherent global scheduling over the solution process.

\newpage
\section{Conclusion}

\textsc{PRL-Bench} is introduced as a research-oriented benchmark to systematically evaluate the capability boundaries of large language models in realistic physics research settings. Unlike prior benchmarks centered on closed-form problems, \textsc{PRL-Bench} emphasizes exploration-oriented task formulation, long-horizon reasoning, and the integration of heterogeneous tools, thereby more faithfully reflecting the structure of authentic scientific inquiry.

Constructed from 100 curated \emph{Physical Review Letters} papers and validated by domain experts, the benchmark spans five major subfields of modern physics and encodes both analytical and computational components within each task. Our evaluation demonstrates that even frontier models achieve limited performance, with the best overall score remaining below 50, revealing a substantial gap between current LLM capabilities and the requirements of autonomous scientific research.

Our analysis further reveals that this gap is not attributable to a single failure mode, but arises from a combination of deficiencies in domain knowledge, derivation stability, numerical reliability, and, critically, long-horizon task adaptation. In particular, the prevalence of conceptual and formulaic errors, together with unstable reasoning trajectories and incomplete solutions, suggests that current models are not robust enough in research planning and long-horizon exploration.

These results highlight that long-horizon reasoning, adaptive methodology selection, and the coordination of multi-step workflows remain fundamental challenges for current systems. \textsc{PRL-Bench} thus provides a rigorous and scalable testbed for future research on AI scientists and long-horizon scientific reasoning.

\section{Limitations and Future Work}
\label{sec:dis}

Despite its design, \textsc{PRL-Bench} involves several limitations. First, compared to authentic research settings, tasks provide relatively richer background information to ensure well-defined objectives and uniquely verifiable answers. While necessary for objective evaluation, this design partially reduces the intrinsic difficulty of open-ended scientific exploration. For the same reason, the benchmark does not explicitly incorporate the process of falsifying incorrect hypotheses, which is a central component of real scientific reasoning.

Second, although all tasks are carefully constructed and cross-validated by domain experts, annotation imperfections may still exist. We plan to continuously refine and expand the benchmark through iterative expert review and community feedback.

Finally, the division into five subfields is inherently approximate. Many research problems—such as quantum many-body systems—naturally span multiple domains, and strict categorization may not fully capture their interdisciplinary nature.

Future work will focus on increasing the openness of task formulation, incorporating elements of hypothesis generation and falsification, and extending the benchmark to cover broader domains and more diverse research paradigms.

\section*{Appendix A: Full Sample Task in \textsc{PRL-Bench}}
\begin{tcolorbox}[title=Sample Task: Tensor-network simulation in (2+1)d lattice gauge theory, colback=gray!5, colframe=gray!75]

\textbf{Introduction}

Traditional Monte Carlo sampling for lattice gauge fields is effective for equilibrium properties, but faces major obstacles for real-time dynamics and sign problems. Tensor-network approaches provide a classical simulation strategy that avoids the sign problem. While (1+1)d systems are well controlled, extending to (2+1)d with PEPS remains challenging due to higher-dimensional entanglement and exact gauge constraints.

We consider an Abelian lattice gauge theory on a square lattice with Hamiltonian
\[
H = H_M + H_B + H_E,
\]
with gauge invariance enforced by Gauss-law constraints. A gauge-invariant PEPS ansatz is adopted, with vertex tensors $A$ and link tensors $B$. Consider open boundary conditions.

\vspace{0.5em}
\textbf{Core tasks}
\vspace{-0.5em}

\begin{enumerate}
\item (2pt) \textbf{Pure $\mathbb{Z}_3$ gauge theory.} 
For a $5\times 5$ lattice with $(D, D_k, h)=(6,2,1)$, characterize the local response of the ground-state energy to the coupling $g$ at $g=0.375$ by determining its first and second derivatives.

\item (3pt) \textbf{Odd $\mathbb{Z}_2$ theory.} 
On a $6\times 6$ lattice with static charges $Q_\mathbf{x}=1$, evaluate the squared expectation values of the staggered observables $D_x$ and $D_y$ at $(h,g)=(1,0.4)$, and assess their symmetry relation.

\item (1pt) \textbf{$\mathbb{Z}_2$ gauge theory with hard-core bosons.} 
For $Q_\mathbf{x}=0$ and $(m,h,g,J)=(0,1,0.33,0.5)$ on a half-filled $4\times 4$ lattice, determine the ground-state expectation value of the Hamiltonian restricted to the central $3\times 3$ bulk.

\item (5pt) \textbf{Real-time dynamics.} 
Starting from the matter-free $\mathbb{Z}_2$ ground state on a $6\times 6$ lattice, create a localized vison excitation and evolve it in real time with $\Delta t=0.005$ at $(g,h)=(0.1,1.0)$. Evaluate:
\begin{itemize}
\item At $T=4.5$, the plaquette $\langle P_{0,0}\rangle$ and total energy $\langle H\rangle$;
\item At $T=6.0$, the plaquette $\langle P_{0,1}\rangle$ and total energy $\langle H\rangle$;
\item At $T=7.5$, the plaquette $\langle P_{1,1}\rangle$ and total energy $\langle H\rangle$.
\end{itemize}
\end{enumerate}

\textbf{Answers \& Rubrics}
\vspace{-0.5em}
\begin{enumerate}
\item Answer: (1pt) $\frac{\partial \langle H\rangle}{\partial g}=29.2609913710$,\ (1pt) $\frac{\partial^2 \langle H\rangle}{\partial g^2}=-179.0345184105$

\item Answer:\ (1pt) $\langle D_x\rangle^2=0.2421075221119777$, \ (1pt) $\langle D_y\rangle^2=0.2421075221119777$

Rubrics: (1pt) Identifies that $C_4$ symmetry enforces $\langle D_x\rangle^2=\langle D_y\rangle^2$.

\item Answer: (1pt) $\langle h_c\rangle=-4.3252682631$ 

\item Answer:
\begin{itemize}
\item (1pt) $\langle P_{0,0}\rangle=0.56388$, \quad (1pt) $\langle H\rangle=-34.64316$
\item (1pt) $\langle P_{0,1}\rangle=0.80091$, \quad (1pt) $\langle H\rangle=-34.64316$
\item (1pt) $\langle P_{1,1}\rangle=0.65918$, \quad (1pt) $\langle H\rangle=-34.64316$
\end{itemize}
Rubrics:
\begin{itemize}
\item (1pt) Recognizes conservation of total energy during real-time evolution.
\item (1pt) Infers vison propagation from spatiotemporal plaquette dynamics.
\end{itemize}

\end{enumerate}

\end{tcolorbox}

\section*{Appendix B: Evaluation Prompt}
\begin{tcolorbox}[
  title=Judge Model Prompt in Evaluation, colback=gray!5, colframe=gray!75]

\ttfamily
\small

You are a strict evaluator for research-oriented physics tasks.

You will be given:
\begin{itemize}
    \item Problem (including subtasks): \{problem\}
    \item Reference answer \& rubrics: \{answer\_and\_rubrics\}
    \item Candidate answer: \{answer\}
\end{itemize}

Evaluate the candidate strictly against the reference answer and rubric, which serve as the sole gold standard.

\textbf{Scoring}
\begin{itemize}
    \item Assign a score normalized to a 0--100 scale.
    \item The reference answer and rubric contain pre-assigned points for both the final answers and intermediate rubric items.
    \item Score the candidate by strictly summing the points of the items it correctly satisfies, according to the predefined allocation.
    \item Credit should be given only for items that are correctly answered and sufficiently matched to the rubric requirements.
\end{itemize}

\textbf{Error types (choose exactly one per subtask)}
\begin{itemize}
    \item Formulaic or conceptual error: inappropriate choice of theoretical models or formulas, incorrect use of boundary conditions, etc.
    \item Derivation error: invalid steps in the reasoning chain, such as introducing unreal formulas or unjustified assumptions.
    \item Calculation error: algebraic or numerical mistakes.
    \item Incomplete: omitted or partial answers, or failure to satisfy task requirements.
    \item Correct: the subtask is essentially correct.
\end{itemize}

\textbf{Requirements}
\begin{itemize}
    \item Perform evaluation at the subtask level (as defined in the problem).
    \item For each subtask:
    \begin{itemize}
        \item determine correctness (true/false),
        \item assign exactly one primary error type,
        \item provide a brief reason (1--3 sentences).
    \end{itemize}
    \item If a subtask is correct, the error type must be Correct.
    \item Be concise and strictly grounded in the rubric.
\end{itemize}

\textbf{Output format}
\begin{verbatim}
SCORE: <0-100, two decimal places>
{
  "subtasks": [
    {"subtask_id": "...", "is_correct": true,
     "primary_error_type": "Correct", "reason": "..."}
  ]
}
\end{verbatim}

\textbf{Constraints}
\begin{itemize}
    \item The number of subtasks must match those in the problem.
    \item primary\_error\_type must be chosen strictly from the defined set.
    \item No additional text or fields beyond the specified format.
\end{itemize}

\end{tcolorbox}

\newpage
\bibliography{main}

\end{document}